\documentclass[10pt,twocolumn,letterpaper]{article}

\newif\ifreview 
\newif\ifarxiv \newcommand{\arxiv}{\arxivtrue}
\newif\ifcamera 
\newif\ifrebuttal 

\arxiv

\ifreview \usepackage[review]{iccv} \fi
\ifarxiv \usepackage[pagenumbers]{iccv} \fi
\ifrebuttal \usepackage[rebuttal]{iccv} \fi
\ifcamera \usepackage{iccv} \fi

\usepackage{lipsum}
\usepackage{booktabs, makecell, multirow}

\usepackage{algorithm}
\usepackage{algpseudocode}
\usepackage{amsmath}
\usepackage{mypsudocode}

\makeatletter
\algrenewcommand\ALG@beginalgorithmic{\footnotesize}
\makeatother

\definecolor{iccvblue}{rgb}{0.21,0.49,0.74}
\usepackage[pagebackref,breaklinks,colorlinks,allcolors=iccvblue]{hyperref}

\def\paperID{1643} 
\def\confName{ICCV}
\def\confYear{2025}

\title{VRsketch2Gaussian: 3D VR Sketch Guided 3D Object Generation with Gaussian Splatting}

\newcommand{\mysubsub}[1]{\noindent\textbf{#1.}}

\author{
Songen Gu$^{1,2*}$ \quad Haoxuan Song$^{2*}$ \quad Binjie Liu$^{3}$ \quad Qian Yu$^{4}$ \quad Sanyi Zhang$^{3}$ \\
Haiyong Jiang$^{2}$ \quad Jin Huang$^{2}$ \quad Feng Tian$^{1,2\dagger}$ \\
$^1$ Institute of Software, CAS \quad
$^2$ UCAS \\
$^3$ Communication University of China \quad 
$^4$  Beihang University \\
}

\begin{document}

\twocolumn[{%
\renewcommand\twocolumn[1][]{#1}%
\maketitle
\centerline{

\includegraphics[width=1\linewidth]{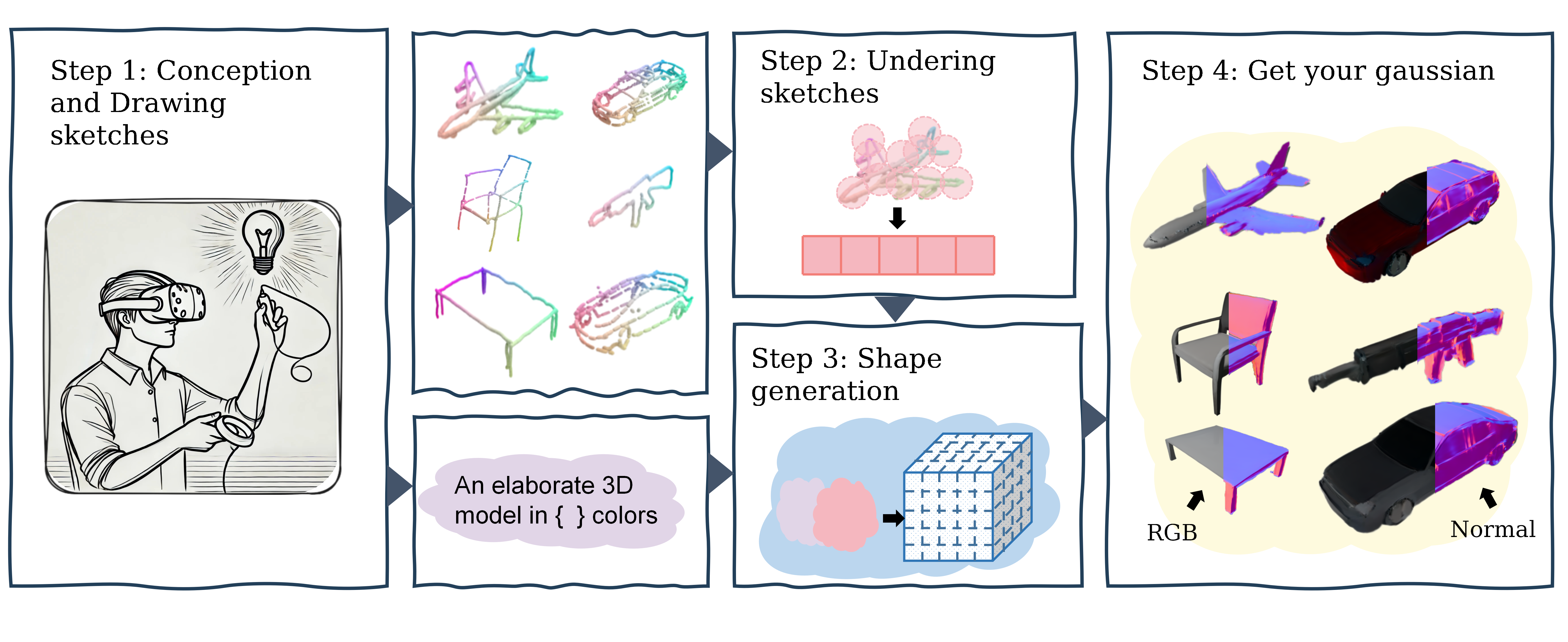}
}
\vspace{-2em}
\captionof{figure}{We present \textbf{VRsketch2Gaussian}, a 3D generation pipeline guided by VR sketches. (1) The user provides a VR sketch and a text prompt. (2) A VR sketch encoder extracts a multi-modal aligned sketch embedding. (3) A 3D-native generation model synthesizes 3D Gaussians guided by the fused features of the VR sketch and text. (4) The final 3D object is represented in the 3D Gaussian.}
\vspace{1em}
\label{fig:teaser}
}]

\ifarxiv
\let\thefootnote\relax\footnotetext{$*$ Equal contribution.}
\let\thefootnote\relax\footnotetext{$\dagger$ Corresponding author.}
\fi

\begin{abstract}

We propose \textbf{VRSketch2Gaussian}, a first VR sketch-guided, multi-modal, native 3D object generation framework that incorporates a 3D Gaussian Splatting representation.
As part of our work, we introduce \textbf{VRSS}, the first large-scale paired dataset containing VR sketches, text, images, and 3DGS, bridging the gap in multi-modal VR sketch-based generation.  
Our approach features the following key innovations:  
(1) \textbf{Sketch-CLIP feature alignment.} We propose a two-stage alignment strategy that bridges the domain gap between sparse VR sketch embeddings and rich CLIP embeddings, facilitating both VR sketch-based retrieval and generation tasks.
(2) \textbf{Fine-Grained multi-modal conditioning.} We disentangle the 3D generation process by using explicit VR sketches for geometric conditioning and text descriptions for appearance control. To facilitate this, we propose a generalizable VR sketch encoder that effectively aligns different modalities.  
(3) \textbf{Efficient and high-fidelity 3D native generation.} Our method leverages a 3D-native generation approach that enables fast and texture-rich 3D object synthesis.  
Experiments conducted on our VRSS dataset demonstrate that our method achieves high-quality, multi-modal VR sketch-based 3D generation.  
We believe our VRSS dataset and VRsketch2Gaussian method will be beneficial for the 3D generation community.

\end{abstract}    
\section{Introduction}


3D Gaussian Splatting\cite{kerbl3DGaussianSplatting2023} has gained popularity due to its explicit Gaussian representation and incredibly fast rendering capabilities. An increasing number of 3D generation works \cite{linDiffSplatRepurposingImage2025,tangLGMLargeMultiView2024} now use it as the representation for 3D generation.
existing text-to-3D methods \cite{longWonder3DSingleImage2024} can generate a 3D model from a text prompt, which is convenient for novices. However, text conditions achieving detailed control over the generated model with text conditions remain challenging, even with elaborate text prompts.


Sketching, on the other hand, is a fundamental tool for expressing ideas and plays a critical role in the design industry, helping artists create 3D prototypes and refine designs \cite{zhou2023ga}. 
With advancements in VR and the Metaverse \cite{dreyVRSketchInExploringDesign2020, jiangHandPainter3DSketching2021}, we recognize that VR sketches serve as an intuitive and easily created tool for designers to express their ideas, offering significant potential for improving 3D generation with fine-grained geometric control.

Hence, VR sketching to 3D Gaussian generation could be highly beneficial, as it provides fine-grained control and enables fast generation. However, this topic has been scarcely explored. We identify several key challenges:
\mysubsub{(1) Limited Paired Sketch to Gaussian Data}
Human-collected VR sketch data is crucial for VR sketch-conditioned generation. However,  existing datasets are limited in size and diversity. Luo \etal \cite{luoFineGrainedVRSketching2022} introduced a VR sketch dataset containing 1,497 3D VR sketch–point cloud pairs, but it only includes ShapeNet chairs, restricting its generalizability for VR sketch-to-3D generation.  
Chen \etal \cite{chenRapid3DModel} later expanded the dataset to include six ShapeNet categories, totaling 4,200 VR sketches. However, this dataset is not publicly available and still lacks diversity in object categories. Moreover, existing datasets provide only paired VR sketches and ground-truth shapes, limiting their utility for multi-modal generation.  
In this work, we propose a high-quality, multi-class, and multi-modal VR sketch dataset called \textbf{VRSS}.  
To overcome the limited category diversity, we provide the dataset of 55 categories with a total of 2097 paired samples. 
To enable multi-modal conditional generation, we provide matched VR sketches, text, and images.  
To support recent advancements in 3D Gaussian Splatting (3DGS) and native 3D generation, we include corresponding point clouds and 3DGS representations.
We believe the VRSS dataset will bridge the gap in VR sketch-based 3D generation, fostering more robust and versatile research in this field.
\mysubsub{(2) The Abstract Nature of Freehand Sketching}
Freehand VR sketches are abstract and not precise, meaning we cannot directly leverage depth information from them. To address this, we adopt contrastive learning to train a sketch encoder that aligns the sketch embedding space with the CLIP embedding space.
Furthermore, VR sketches primarily convey geometry information, so many VR sketch-based modeling methods are limited to generating only colorless shapes. To overcome these limitations, we present the first multi-modal 3D native object generation framework, \textbf{VRSketch2Gaussian}, which is conditioned on both VR sketches and text.

\mysubsub{(3) Poor 3D Gaussian Geometry in Generaion}
3D Gaussian representations tend to suffer from low geometric fidelity, as they describe the scene using discrete primitives. This issue is exacerbated when the generation model directly optimizes the Gaussian parameters using MSE loss. To alleviate this, we employ a Perceiver-based \cite{jaegle2021perceiver} reducer to preserve detailed geometry from the sketches. In addition, we apply a joint refinement strategy for appearance and geometry during training, which enhances the final generated results.

Both qualitative and quantitative evaluations conducted on the FVRS and VRSS datasets demonstrate that our method achieves high-quality VR sketch-guided 3D object generation.

In conclusion, our contributions are listed below:
\begin{itemize}
    \item We introduce a new dataset containing VR sketches, text, images, and 3D Gaussian, bridging the gap in multi-modal VR sketch-based 3D generation.
    \item We propose a contrast learning-based sketch to image and point cloud alignment, enabling extract representative VR sketch embedding.
    \item We adopt a 3D native generation method that integrates both sketch and text conditions, enabling the generation of high-quality 3D Gaussian models with fine geometric and visual fidelity.

\end{itemize}

\section{Related Work}

\subsection{Sketch-Based Shape Modeling and Generation}

\mysubsub{2D Sketch-Based Modeling} 2D sketch-based modeling method takes either a single-view image \cite{zhangSketch2ModelViewAware3D2021} or multi-view sketch images \cite{zhouGASketchingShapeModeling2023} as input and produces a 3D shape.
Early approaches relied on shape retrieval techniques, which aim to extract discriminative features from a sketch and retrieve the most similar 3D shape from a database. However, this method is inherently limited by the size of the shape database.
Modern methods \cite{guillard2021sketch2mesh,zhangSketch2ModelViewAware3D2021,gaoSketchSamplerSketchbased3D2022,zhengLocallyAttentionalSDF2023,bandyopadhyay2024doodle,liuSketchDreamSketchbasedTextto3D2024,wang3DShapeReconstruction2022} leverage single-view sketch images to generate 3D shapes; however, these approaches often suffer from ambiguities due to occlusion in the single-view sketches.
Other methods \cite{delanoy3DSketchingUsing2018,delanoyCombiningVoxelNormal2019,zhou2023ga,zhongPracticalSketchBased3D2021a} attempt to alleviate this ambiguity by incorporating multi-view sketches. Nonetheless, they still face challenges related to inconsistent results.

\mysubsub{VR Sketch-Based Modeling}
Recent advancements in virtual reality (VR) have led to the development of a wide range of 3D creation tools \cite{brushOpenBrush,GravitySketch3D,VRSketchDesign}. VR offers an immersive creative environment and introduces new input modalities for sketch-based modeling.
Traditional methods either focus on enhancing user experience and improving usability \cite{yuScaffoldSketchAccurateIndustrial2021,2018SymbiosisSketchC,rosalesStripBrushConstraintRelaxed3D2021} or aim to convert sketches into well-organized geometric representations, such as curve networks \cite{yuCASSIECurveSurface2021} or structured 3D surfaces \cite{yuPiecewisesmoothSurfaceFitting2022,rosalesSurfaceBrushVirtualReality2019}.
generation-based methods guided by VR sketches are less likely to be exported.
Luo et al. \cite{luo3DVRSketch2024} utilizes a 3D shape generation network conditioned on 3D VR sketches, aiming to reconstruct geometrically realistic 3D shapes from sketches created by novices. This technique addresses sketch ambiguity by generating multiple 3D shapes that align with the original sketch’s structure. It leverages multi-modal 3D shape representations and a dedicated loss function to improve fidelity to the input sketches.
Deep3DVRSketch \cite{chenRapid3DModel} takes 3D sketches and employs a diffusion model to generate consistent 3D models. It uses a coarse-to-fine pipeline, starting with an occupancy diffusion to create a low-resolution voxel and refining it into a high-resolution SDF voxel with SDF-diffusion \cite{zhengLocallyAttentionalSDF2023}. For sketch conditioning, it employs an off-shelf point-cloud encoder \cite{zhouUni3DExploringUnified2023b} to obtain the latent representation and aligns it with CLIP’s \cite{c57293882b2561e1ba03017902df9fc2f289dea2} image latent space.
These methods have pioneered VR-based 3D shape generation; however, they still face challenges, such as the inability to generate colorful 3D objects, limited conditioning classes and modalities, and complex pipelines.

\subsection{3D Object Generation}

\mysubsub{2D Diffusion Distillation-Based} Due to the success of 2D diffusion models \cite{rombachHighResolutionImageSynthesis2022}, several works have attempted to leverage 2D image-based diffusion models for 3D generation. SDS \cite{4c94d04afa4309ec2f06bdd0fe3781f91461b362} and SJC \cite{wangScoreJacobianChaining2023} are pioneering text-to-image diffusion models for generating 3D objects, inspiring subsequent text-to-3D and image-to-3D approaches \cite{hong20243dtopia,linMagic3dHighresolutionTextto3d2023a,tangDreamgaussianGenerativeGaussian2023a,yuPointsto3dBridgingGap2023,chenIt3dImprovedTextto3d2024}. These methods typically employ a 3D representation (e.g., NeRF, 3DSG, or Mesh) to render multi-view images and utilize diffusion priors to optimize the 3D structure. Despite significant advancements in 3D generation, these approaches still suffer from multi-face artifacts and inconsistencies due to the lack of direct 3D supervision.

\mysubsub{Multi-View Reconstruction-Based}
To address the limitations caused by SDS, recent methods \cite{shiMvdreamMultiviewDiffusion2023,wangImageDreamImagePromptMultiview2023,voletiSV3DNovelMultiview2024,liuSyncdreamerGeneratingMultiviewconsistent2023,liInstant3DFastTextto3D2023,tangLGMLargeMultiView2024} utilize multi-view or video diffusion models to generate consistent multi-view images in a single step, followed by sparse view reconstruction to obtain 3D models. However, these approaches typically involve a two-stage inference process—multi-view generation followed by reconstruction—which limits their inference efficiency.

\mysubsub{Native 3D-based}
Native 3D generation methods aim to directly produce 3D representations, such as Voxel \cite{renXcubeLargescale3d2024}, SDF \cite{parkDeepSDFLearningContinuous2019a, chengSdfusionMultimodal3d2023}, mesh \cite{siddiquiMeshgptGeneratingTriangle2024, liCraftsManHighfidelityMesh2024}, and point cloud \cite{nicholPointESystemGenerating2022}. Recent advancements explore new representations like 3DGS \cite{linDiffSplatRepurposingImage2025, zhangGaussianCubeStructuredExplicit2024} and triplane \cite{caoLargeVocabulary3DDiffusion2023}. Despite their success, these methods still face challenges such as high training costs and complex data preparation.


\section{Method}
\begin{figure*}[htbp]
        \centering
    \includegraphics[width=1\linewidth]{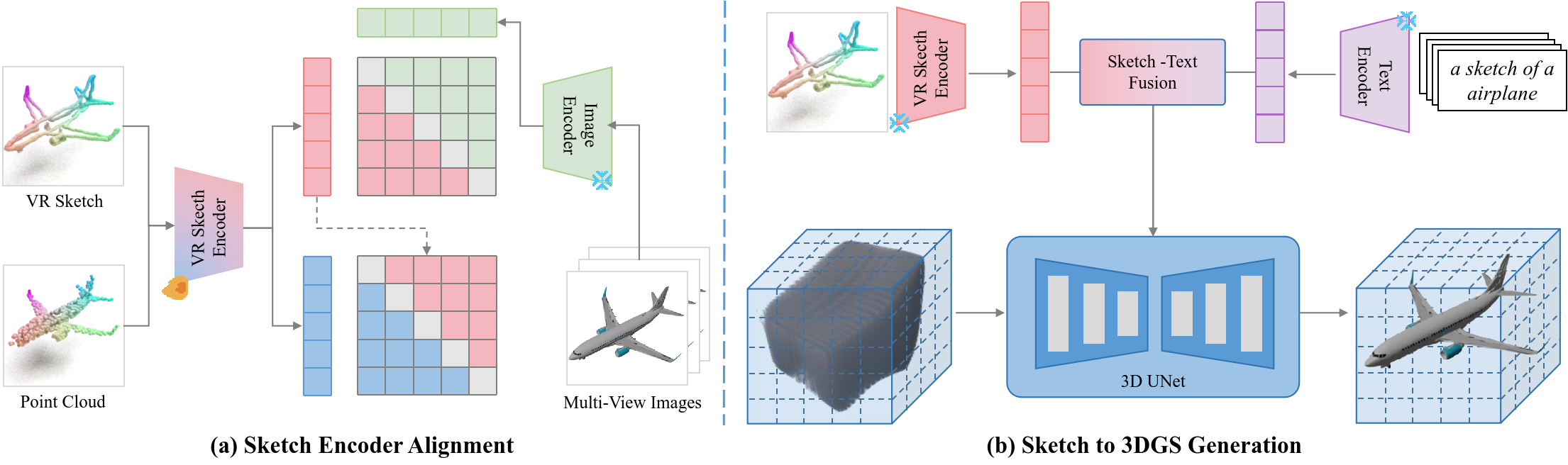}
    \caption{\textbf{Pipeline.} Our method consists of two stages: (a) We first train a VR sketch encoder that aligns with the CLIP embedding space using contrastive learning, and (b) we use both sketch and text for multi-modal conditional generation.}
    \label{fig:enter-label}
\end{figure*}

\subsection{Datase Collection}
We introduce \textbf{VRSS} (short for VR Sketch to Splat), the first dataset that pairs VR sketches with text, images, point clouds, and 3D Gaussian.  
Our VR sketch collection procedure follows \cite{luoFineGrainedVRSketching2022}, where participants wear a VR headset and redraw a given 3D shape as a VR sketch. However, unlike \cite{luoFineGrainedVRSketching2022}, we observed that users often introduce discrepancies (random stretch, scale, and offset errors) when the 3D ground truth shape is not placed in a consistent reference frame.  
To address this issue, we modified our data collection pipeline by ensuring that participants sketch within the same coordinate system as the 3D shape. Additionally, we implemented a \textbf{shape switch button}, requiring participants to finalize their sketch only when toggling off the reference shape does not cause significant visual differences. This ensures that no critical shape details are omitted, resulting in more precisely aligned and high-quality VR sketches.  
We employ hierarchical sampling to select 2097 shapes from the original ShapeNet dataset, covering 55 object categories.
Text prompts are obtained following \cite{liuOpenShapeScaling3D2023b}. We use Blender to reshape view-dependent images. For each shape, we render 72 RGB images from random viewpoints. Depth and Normal are also rendered alongside. These RGB images are later trained to get 3DGS. Point cloud are uniform sampled from the original 3D shape in ShapeNet.

\subsection{VR Sketch Encoder}

The VR sketch encoder is a fundamental component of our framework. Learning a compact and representative sketch feature is crucial for sketch conditioned generation, as it directly impacts the fidelity of the synthesized 3D shapes. Since CLIP features are abstract and well-aligned, we aim to align the sketch encoder features with CLIP space. To bridge the domain gap between sparse sketch inputs and rich CLIP embeddings, we propose a two-stage alignment strategy that maps both 3D shapes and freehand VR sketches into the unified CLIP space. This alignment enables cross-modal semantic consistency while preserving the geometric priors learned from large-scale shape datasets.

\mysubsub{Sketch-Shape-CLIP alignment} We adopt Point-BERT \cite{yu2021pointbert} as the backbone, addressing the challenge of mapping sparse sketch point clouds to CLIP space under limited sketch data. Despite sharing the same data format with standard point clouds, VR sketches exhibit distinct characteristics: 1) sparser point distributions reflecting human drawing habits; 2) only show the most basic features of the shape. This difference makes us treat sketches unlike prior works \cite{liuOpenshapeScaling3d2023, zhouUni3DExploringUnified2023b}. We decompose the training procedure into two steps: 1) Initialize the encoder on ShapeNet\cite{changShapeNetInformationRich3D2015} point clouds to inherit CLIP-aligned shape semantics. This establishes a geometry-aware foundation by leveraging the dense shape-text-image correlations; 2) fine-tune using our collected VR sketch dataset to capture human drawing patterns. To enhance geometric sensitivity and visual alignment.

\mysubsub{Step 1: Shape-Only Training} In this step, we aim to align the shape point cloud to CLIP space. We follow the previous point cloud contrastive learning paradigm and calculate the point cloud-text-image contrast loss. The training  objectives are as follows:
Given a  triplet $\left\{\left(P_i,T_i, I_i\right)\right\}$, $P_i$ is the corresponding 3D shape point cloud,  $T_i$ is the text, and $I_i$ is the rendered image. The contrastive loss between point clouds and images is denoted as:
\begin{equation}
\begin{aligned}
\ell^{P \to I} =  \sum_i^{N} & \log \frac{\exp \left(h_i^P \cdot h_i^I / \tau\right)}{\sum_j \exp \left(h_i^P \cdot h_j^I / \tau\right)}
\end{aligned}
\end{equation}

\begin{equation}
\begin{aligned}
\ell^{I \to P} =  \sum_i^{N} & \log \frac{\exp \left(h_i^I \cdot h_i^P / \tau\right)}{\sum_j \exp \left(h_i^I \cdot h_j^P / \tau\right)}
\end{aligned}
\end{equation}

\begin{equation}
\mathcal{L}_{P \leftrightarrow I} = -\frac{1}{2} \Big(\ell_{P \to I} + \ell_{I \to P}\Big)
\end{equation}
where $ h_i^{p} $ is point cloud features and  $h_i^{I} $ is image features.$ \tau\ $ is the temperature parameter.

In this case,the overall objective becomes:

\begin{equation}
\begin{aligned}
\mathcal{L}_{\text{All}} = 
\mathcal{L}_{P \leftrightarrow I} + 
\mathcal{L}_{P \leftrightarrow T}
\end{aligned}
\end{equation}

where $\mathcal{L}_{P \leftrightarrow T}$ is similar to Equation,$h_i^{T} $ is the text feature. In this step, our encoder can learn rich geometric prior knowledge from the data.

\mysubsub{Steps 2: Sketch-Shape Alignment}  Here we perform using the sketch data to train the encoder aligned the sketch to shape. Given a  triplet $\left\{\left(S_i,P_i, I_i\right)\right\}$, $S_i$ is the VR sketch, $P_i$ is point cloud and $I_i$ is image. First, we aligned the sketch and shape to CLIP image feature. With 3D representation \(h^{p}, h^{s}\) and image features \(h^{I}\), the contrastive loss between sketches and images is denoted as:
\begin{equation}
\mathcal{L}_{S \leftrightarrow I} = -\frac{1}{2} \Big(\ell_{S \to I} + \ell_{I \to S}\Big)
\end{equation}
Besides that \(\mathcal{L}_{P \leftrightarrow S}\) and \(\mathcal{L}_{P \leftrightarrow I}\) are formulated similarly to the conventional contrastive term. 

Beyond these contrastive losses, we employ a triplet loss term to explicitly ensure that the feature of sketch (\(h^s\)) is closer to its corresponding shape feature (\(h^p\)) than to any other shapes:
\begin{equation}
\mathcal{L}_{\text{T}} = \frac{1}{n} \sum_i \max \Big( 0, \| h_i^s - h_i^p \|_2 - \min_{j \neq i} \| h_i^s - h_j^p \|_2 + m \Big)
\end{equation}
where \(m\) is the margin hyperparameter.

Incorporating all these terms, the overall loss function is defined as:
\begin{equation}
\begin{aligned}
\mathcal{L}_{\text{All}} = \frac{1}{3}  \Big(
\mathcal{L}_{P \leftrightarrow I} + 
\mathcal{L}_{S \leftrightarrow I} + 
\mathcal{L}_{P \leftrightarrow S} 
\Big)
+ \mathcal{L}_{\text{Triplet}}
\end{aligned}
\end{equation}

Note that in order to ensure the alignment of point cloud and CLIP feature, we still use $\mathcal{L}_{P \leftrightarrow I}$ for alignment in the fine-tuning stage, and align the sketch with the image and point cloud. In addition, to improve the model's ability to understand different sketches of the same category, we introduced triplet loss to improve the model's ability to understand the sketch in detail.

\subsection{Sketch-to-3DGS Generaion}


\begin{algorithm}[htbp]
    \caption{Efficient Constrained Densification}
    \label{alg:optimization}
    \begin{algorithmic}[1]

        \Statex\textbf{Input:} Pretrained variable number of Gaussians, maximum number of Gaussians $G_{nf}$, maximum number of iterations $N_{max}$
        
        \Statex\textbf{Output:} Fixed number of Gaussians $G_f$
        
        \If {$\text{len}(G_{nf}) \le N_{max} $} \Comment{Load Pretrained Gaussians}
        \State $M, S, C, A \gets G_{nf}$ 
        \Else								
        \State   $M, S, C, A \gets$ FarthestGaussianSample($G_{nf}$, $0.8 \times N_{max}$) \Comment{Sample Gaussians based on farthest points}
        \EndIf
        \State $i \gets 0$	\Comment{Iteration Count}
        
        \While{not converged}
        
        \State $V, \hat{I} \gets$ SampleTrainingView()	\Comment{Camera $V$ and Image}
        \State $I \gets$ Rasterize($M$, $S$, $C$, $A$, $V$)
        
        \State $L \gets Loss(I, \hat{I}) $ \Comment{Loss}
        
        \State $M$, $S$, $C$, $A$ $\gets$ Adam($\nabla L$) \Comment{Backprop \& Step} 

        \State $\Delta N \gets \max\left(N_{max} - \text{len}(M),0\right)$ \Comment{Number of new Gaussians}
        \State $idx_{\text{topK}} \gets$ TopK($\nabla_p L$, $\Delta N$) \Comment{Get top $\Delta N$ Gaussians with largest gradients}
        
        \State 
        \If{IsRefinementIteration($i$)}
        \ForAll{$i$,Gaussians $(\mu, \Sigma, c, \alpha)$ $\textbf{in}$ $(M, S, C, A)$}
        \If{$\alpha < \epsilon$ or IsTooLarge($\mu, \Sigma)$}	\Comment{Pruning}
        \State RemoveGaussian()	
        \EndIf
        \If{$\nabla_p L > \tau_p$ and $i \in idx_{\text{topK}}$ } \Comment{Constrained-Densification}
        \If{$\|S\| > \tau_S$}	\Comment{Over-reconstruction}
        \State SplitGaussian($\mu, \Sigma, c, \alpha$)
        \Else								\Comment{Under-reconstruction}
        \State CloneGaussian($\mu, \Sigma, c, \alpha$)
        \EndIf	
        \EndIf
        \EndFor		
        \EndIf
        \State $i \gets i+1$
        \EndWhile
    \end{algorithmic}
\end{algorithm}

\mysubsub{Preliminaries of 3DGS}
3D Gaussian Splatting \cite{kerbl3DGaussianSplatting2023} is a novel explicit 3D representation. It utilizes a bunch of discrete 3D Gaussians as scene elements, which are defined with a center point \(\mu \in \mathbb{R}^3\), a Gaussian covariance matrix \(\Sigma \in \mathbb{R}^{4\times4}\), opacity \(\alpha\), and a view-dependent color feature \(c \in \mathbb{R}\). These properties are all learnable parameters and are optimized through differentiable rasterization. During rasterization, Gaussians are projected onto the image plane as 2D Gaussians and sorted by depth. The final pixel color is computed using alpha blending:
\begin{equation}
C=\sum_{i \in \mathcal{N}} c_i \alpha_i \prod_{j=1}^{i-1}\left(1-\alpha_j\right)
\end{equation}
where $c_i$ is the color and $\alpha_i$ is a multification of opacity and 2D gaussian covariance.
The 3D Gaussian parameters of Gaussian are optimized using a photometric loss.

\mysubsub{Efficient Constrained Densification}
The original 3D Gaussian Splatting training in \cite{kerbl3DGaussianSplatting2023} produces a variable number of Gaussians, which cannot be handled by 3D diffusion models that usually cooperate with voxel data.
\cite{zhangGaussianCubeStructuredExplicit2024} solves this issue with densification-constrained techniques.
However, this technique requires training Gaussian from scratch, resulting in a long preparation time for large-scale Gaussian data.
We notice that we can shorten the training time by converting trained variable number Gaussian to fixed Gaussian. Directly removing or adding Gaussians will corrupt the visual result; hence, we propose EDC (Efficient Constrained Densification) training strategy. Given a pretrained, variable number of Gaussians, we first sample Gaussians less than $N_{\max}$, and we only apply densification to Gaussians with large gradients while ensuring the total number of Gaussians does not exceed $N_{\max}$. This enables fast adaptation from a variable number to a fixed number of Gaussians.

\mysubsub{3D Diffusion for Gaussian Generation}
A 3D diffusion module is utilized for Gaussian generation. To cooperate Gaussian with 3D diffusion, which operates on 3D voxels, a mapping operation using optimal transport is employed \cite{zhangGaussianCubeStructuredExplicit2024}. This structure organizes the Gaussian into a feature voxel grid, denoted as \( v \in \mathbb{R}^{n \times n \times n \times C} \), where \( n^3 = N_{\max} \) and \( C = 14 \), representing the concatenated Gaussian parameters (position, scaling, rotation, opacity, and color).

A 3D U-Net, \( \epsilon_{\theta} \), is applied as a denoiser. During training, Gaussian noise is added to the clean feature voxel grid, and the 3D U-Net learns to predict the noise for a given timestamp \( t \).
The 3D U-Net is optimized with diffusion loss:
\begin{equation}
\mathcal{L}_{diff}=\mathbb{E}_{z, \epsilon \sim n(0,1), t}\left[\left\|\epsilon-\epsilon_\theta\left(z_t, t,h^s,h^t\right)\right\|^2\right]
\end{equation}

For the inference stage, it gradually transforms the noisy feature grid into a clean Gaussian feature.

\mysubsub{Fine-grained Text-Sketch Fusion Conditioning}
VR sketches play a crucial role in representing the geometric intuitive conditions in a 3D generation.
After training the VR sketch encoder, we extract the VR sketch's embedding to serve as the generation condition.
It might seem intuitive to use the output of the last layer of the VR sketch encoder, which produces a reduced embedding with a shape of \( B \times C \). This embedding is aligned with the CLIP embedding and is considered an abstract and representative feature of the input VR sketch. However, we find that simply using this embedding as a condition leads to a loss of detailed information
in the generation process, as the reduction operation discards the spatial information contained in the original VR sketch. An alternative approach is to use the embedding before reduction, which has the shape \( B \times L \times C \), but this is not ideal, as it includes more tokens, thus increasing the computational cost.
To capture the fine-grained spatial information of the VR sketch while reducing the embedding dimension, we utilize a Perceiver-based reducer \cite{jaeglePerceiverGeneralPerception2021}. Specifically, a low-dimensional (16 in our case) learnable query is iteratively used to distill the high-dimensional input sketch embedding (\( L = 129 \) in our case) through asymmetric attention. After this operation, we obtain a low-dimensional, feature-rich sketch embedding $\mathcal{F}_{\text {sketch }}$.

In addition to VR sketch embedding, we also require text to represent abstract and semantic conditions, such as texture and color. Consistent with common practices in text-to-3D generation, we utilize the CLIP text encoder for text feature extraction, the extracted feature is denoted as $h^T$.

We combine the sketch embedding and the text embedding through a concatenation operation to obtain the fused embedding $\mathcal{F}_{\text{fused}} = \text{concat}\left([h^S, h^T]\right)$, which is then fed into the cross-attention module of the 3D U-Net.

\mysubsub{Appearence Geomery joint Refinement}
Directly generating the Gaussain feature though generates well gaussian feature Directly generating the Gaussian feature yields a fine Gaussian feature but poor in novel view rendering. 
To address this issue, a photometric loss is adopted, following \cite{zhangGaussianCubeStructuredExplicit2024}. Specifically, during diffusion training, given a noisy sample $x_t$, the 3D U-Net predicts a noise $\mathbf{z}_\theta\left(\mathbf{x}_t, t\right)$. Using the reparameterization trick, we obtain:
\begin{equation}
\begin{aligned}
\hat{\mathbf{x}}_0=\frac{1}{\sqrt{{\alpha}_t}}\left(
\mathbf{x}_t-\sqrt{1-{\alpha}_t} \mathbf{\mathbf{\epsilon}_\theta} \right)
\end{aligned}
\end{equation}
This approach allows for obtaining a clean sample without performing multi-step denoising. With the clean Gaussian sample, we use it for differentiable rasterization to obtain the predicted RGB image, $I_{pred}$. The Pixel L1 loss and perceptual loss are then applied:
\begin{equation}
\begin{aligned}
\mathcal{L}_{\text {image }}&=\mathbb{E}_{I_{\text {pred }}}\left(\left\|V\left(I_{\mathrm{pred}}\right)-V\left(I_{\mathrm{gt}}\right)\right\|_2^2\right) \\
&+\mathbb{E}_{I_{\mathrm{pred}}}\left(\left\|I_{\mathrm{pred}}-I_{\mathrm{gt}}\right\|_1\right),
\end{aligned}
\end{equation}
Where $V$ is the  VGG-16 network.
Similarly, we perform geometry refinement using the depth and normal of $\hat{\mathbf{x}}_0$. Both depth and normal are rendered with the differentiable rasterizer:
\begin{equation}
\begin{aligned}
        D&=\sum_{i \in \mathcal{N}} d_i \alpha_i \prod_{j=1}^{i-1}\left(1-\alpha_j\right) \\
         N&=\sum_{i \in \mathcal{N}} n_i \alpha_i \prod_{j=1}^{i-1}\left(1-\alpha_j\right) \\
\end{aligned}
\end{equation}
where $d_i$ is the depth from the viewpoint to the Gaussian center, and $n_i$ is the normal of the Gaussian in world coordinates, determined by the shortest axis direction of the covariance matrix. The rendered depth and normal are optimized with the ground truth  depth and normal using L1 loss:
\begin{equation}
\begin{aligned}
  \mathcal{L}_{\text {depth }}&=\mathbb{E}_{D_{\mathrm{pred}}}\left(\left\| D_{\mathrm{pred}}-D_{\mathrm{gt}}\right\|_1\right ) \\
\mathcal{L}_{\text {normal }}&=\mathbb{E}_{N_{\mathrm{pred}}}\left( \left\| N_{\mathrm{pred}}-N_{\mathrm{gt}}\right\|_1 \right) 
\end{aligned}
\end{equation}

Our final training loss is the weighted sum of the diffusion loss and the refinement loss, where $\lambda_1$, $\lambda_2$, $\lambda_3$, and $\lambda_4$ are the weights that balance these losses:

\begin{equation}
\mathcal{L}_{final}=
\lambda_1\mathcal{L}_{diff}+
\lambda_2\mathcal{L}_{\text {image }}+    \lambda_3\mathcal{L}_{\text {depth }}+\lambda_4\mathcal{L}_{\text {normal }}
\end{equation}



\section{Experiments}

\begin{figure*}
    \centering
    \includegraphics[width=0.9\linewidth]{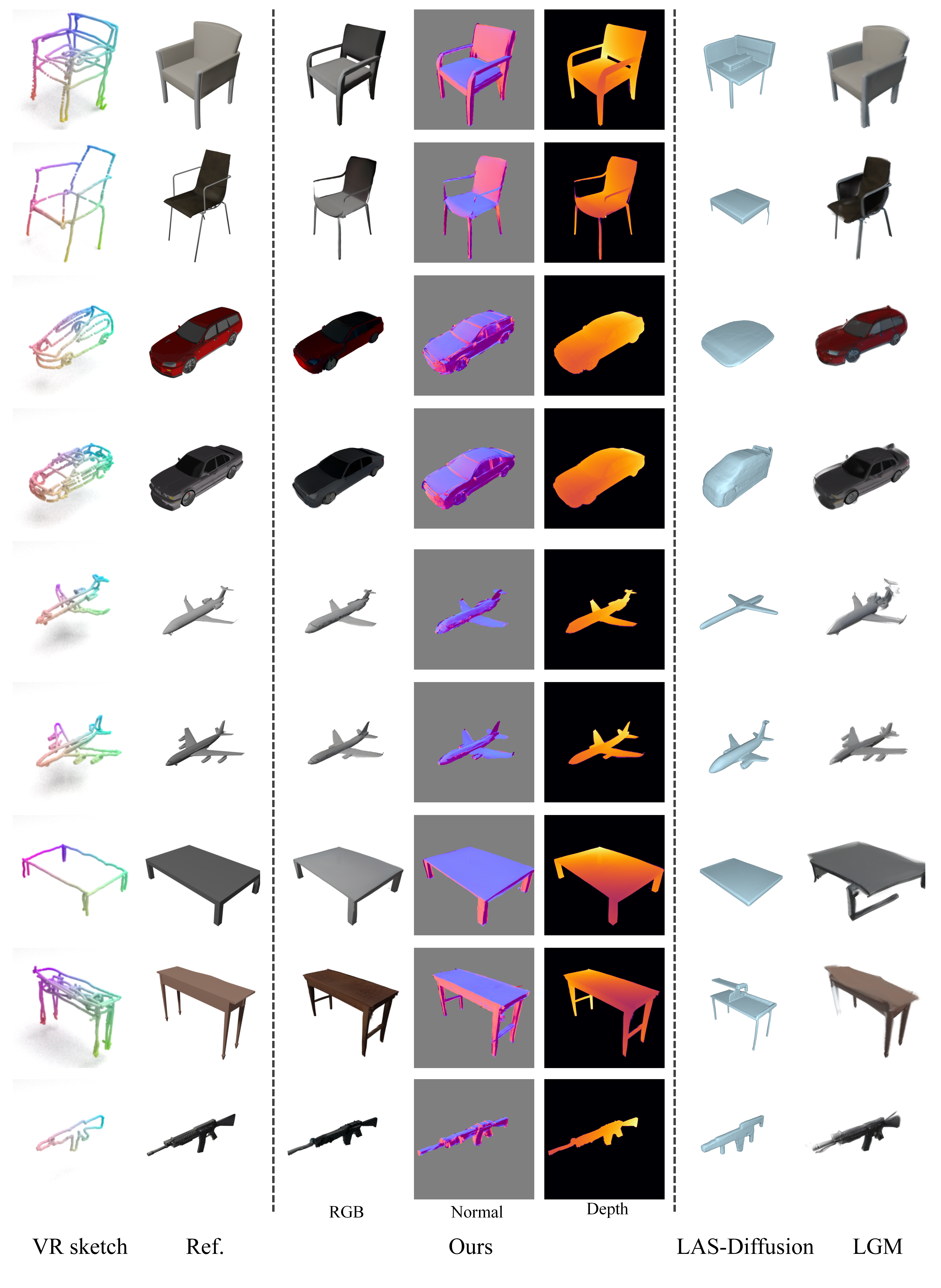}
    \caption{Our Qualitative Results.}
    \label{fig:render_cmp}
\end{figure*}

\subsection{Implementation Details}
\mysubsub{Datasets} 
In the first stage alignment, We evaluate our method on both the VRSS dataset and the public FVRS dataset \cite{luoFineGrainedVRSketching2022}, and we use VRSS in our generation model training. 
FVRS is a human-drawn VR sketch dataset comprising 1,497 3D VR sketches and 3D-shaped pairs of chairs.
Our VRSS dataset is a multi-modal, multi-class sketch dataset containing 55 classes, a total of 2097 paired samples with human hand-collected aligned sketches in VR. We Split the VRSS dataset into train and test, which contains 1678 in training and 419 in testing, respectively.    
\mysubsub{Metrics}.  To evaluate the quality of the sketch-guided 3D generation, we use Chamfer Distance (CD) for geometric evaluation, and FID and CLIP score for evaluating the rendering visual quality. During evaluation, the model is hand-aligned to the same coordinate frame as the ground truth.

\mysubsub{Network} 
Our model adopts an asymmetric U-Net structure consisting of four down blocks, one middle block, and four up blocks. The input voxel resolution is $32\times 32\times 32$, with a channel size of $14$, resulting in a total of $32,768$ output Gaussians. The feature channels for the down, middle, and up blocks are $[128, 256, 384, 512]$, $[512]$, and $[512, 384, 256, 128]$, respectively. Each block comprises residual layers, incorporating downsampling or upsampling layers where applicable. We employ SiLU activation and group normalization to enhance stability.

\mysubsub{Training} 
We use AdamW as our optimizer during the second training stage and fix the learning rate to $1e^{-4}$ for 10,000 iterations. The batch size is set to 80. An EMA technique is applied to smooth the U-Net's parameters. During the first 4,000 iterations, we only supervise the diffusion loss. After 4,000 iterations, in addition to the diffusion loss, we include the rendering loss, depth loss, and normal loss. We conducted our experiment using 8 NVIDIA A8000 graphics cards and trained the model for approximately two days.


\subsection{Sketch Multi-Modal Alignment}
\begin{figure}
    \centering
    \includegraphics[width=0.8\linewidth]{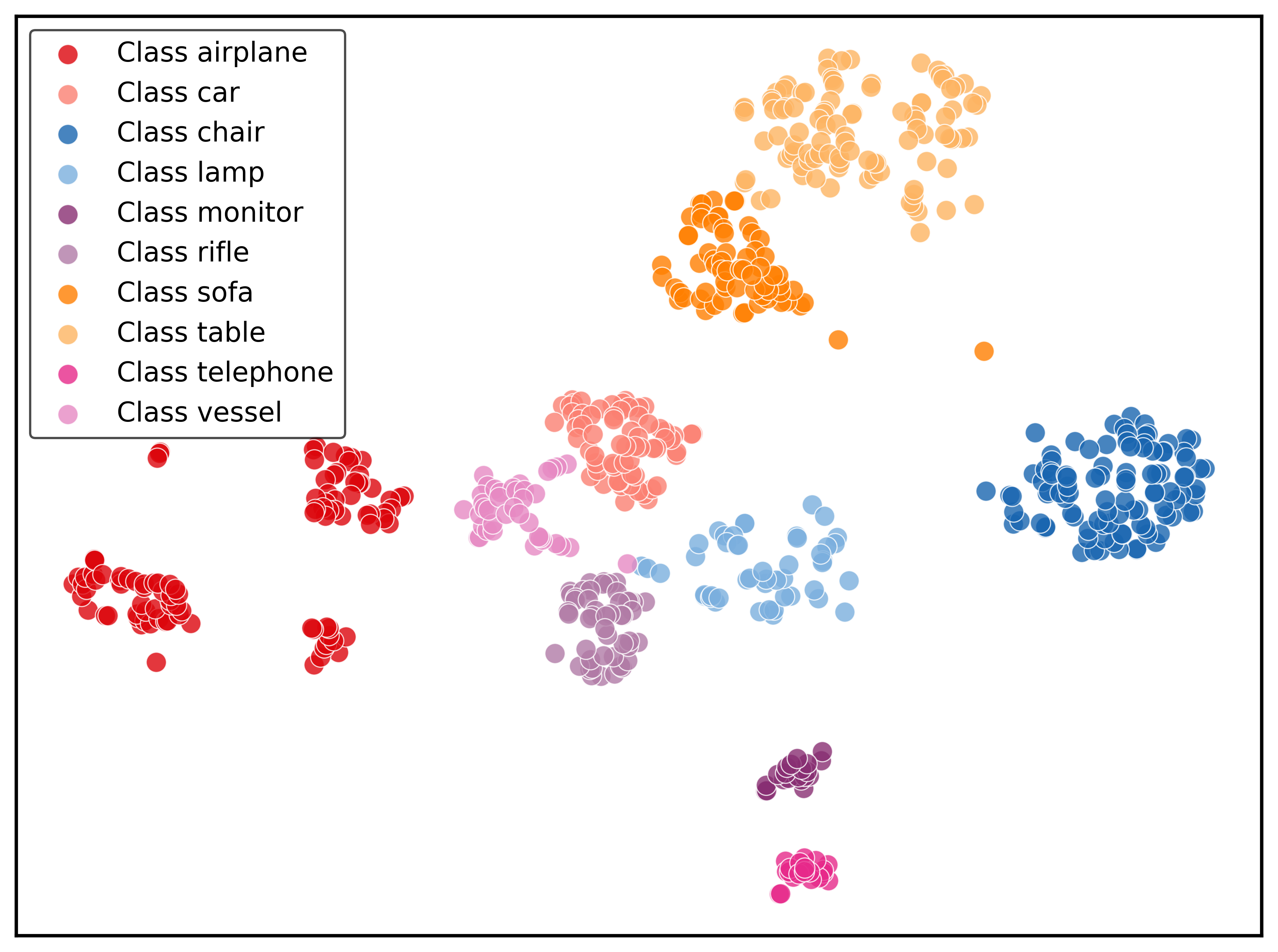}
    \caption{Sketch embedding distribution across categories}
    \label{fig:tsne}
\end{figure}

We evaluate the feature quality of our sketch encoder by performing cross-modal embedding retrieval and classification. 
 We first assess sketch-to-point cloud shape retrieval.
The results are shown in \cref{tab:fvrs}. Our method achieves 32.18\% top-1, 53.80\%, and 62.54\% retrieval accuracy, respectively, achieving state-of-the-art performance on the FVRS dataset.
We also demonstrate sketch-to-image classification. The results are shown in \cref{tab:mm_retival}. For the baseline method, we use the OpenShape\cite{liuOpenShapeScaling3D2023b} point cloud encoder. We apply farthest point sampling to randomly sample points and send them to the OpenShape point cloud encoder.
The results show that our method outperforms the baseline in sketch-to-image classification, achieving a significantly better top-1 accuracy. This demonstrates that our method aligns sketch embeddings well with the CLIP space, which is beneficial for sketch-based cross-modal tasks.
We further demonstrate the distribution of our sketch embedding space by visualizing it with t-SNE. The results shown in \cref{fig:tsne} indicate that most categories in the feature space are separable, suggesting that the sketch embedding is highly representative.

\begin{table}[htbp]
    \centering
   \resizebox{0.8\linewidth}{!}{
    \begin{tabular}{l|cccc}
        \toprule
        \multirow{2}{*}{Method}   
        & \multicolumn{3}{c}{Accuracy $\uparrow $}\\
        \cmidrule(lr){2-4}
        
        & Top-1 & Top-5 & Top-10 \\
        \midrule
        Luo \cite{luoFineGrainedVRSketching2022} & 24.55   & 43.76   & 52.57   \\
        Luo \cite{luoStructureAware3DVR2022a}  &25.45   & 43.96   & 53.37    \\
        Ours (w/o pre-train)  & 29.70 & 48.02 & 55.45 \\
        Ours  & \textbf{32.18} & \textbf{53.80} & \textbf{62.54} \\
        \bottomrule
    \end{tabular}
   }
    \caption{Performance comparison of different methods in the FVRS dataset \cite{luoFineGrainedVRSketching2022}.
    }
    \label{tab:fvrs}
\end{table}

        

\begin{table}[htbp]
    \centering
    \resizebox{0.8\linewidth}{!}{
    \begin{tabular}{l|ccc}
        \toprule
        \multirow{2}{*}{Method} &  
        \multicolumn{3}{c}{Image Acc. $\uparrow $} \\
        \cmidrule(lr){2-4}
        
       & Top-1 & Top-5 & Top-10  \\
        \midrule
        OpenShape \cite{liuOpenShapeScaling3D2023b} & 78.83&	87.70&	90.37      \\
        Ours &   \textbf{89.98}&	\textbf{98.09}&	\textbf{98.57}\\
        \bottomrule
    \end{tabular}
    }
    \caption{Performance comparison of different VR sketch-to-image classification methods in the VRSS datasets.}
    \label{tab:mm_retival}
\end{table}

\subsection{Qualitative and Quantitative Result}

\mysubsub{3D Generation} We compare our 3D generation results with two 3D generation baseline: the 2D sketch-conditioned 3D generation method LAS-Diffusion \cite{zhengLocallyAttentionalSDF2023} and the image-based 3D generation method LGM \cite{tangLGMLargeMultiView2024}.

\begin{figure}[thbp]
    \centering
    \includegraphics[width=1\linewidth]{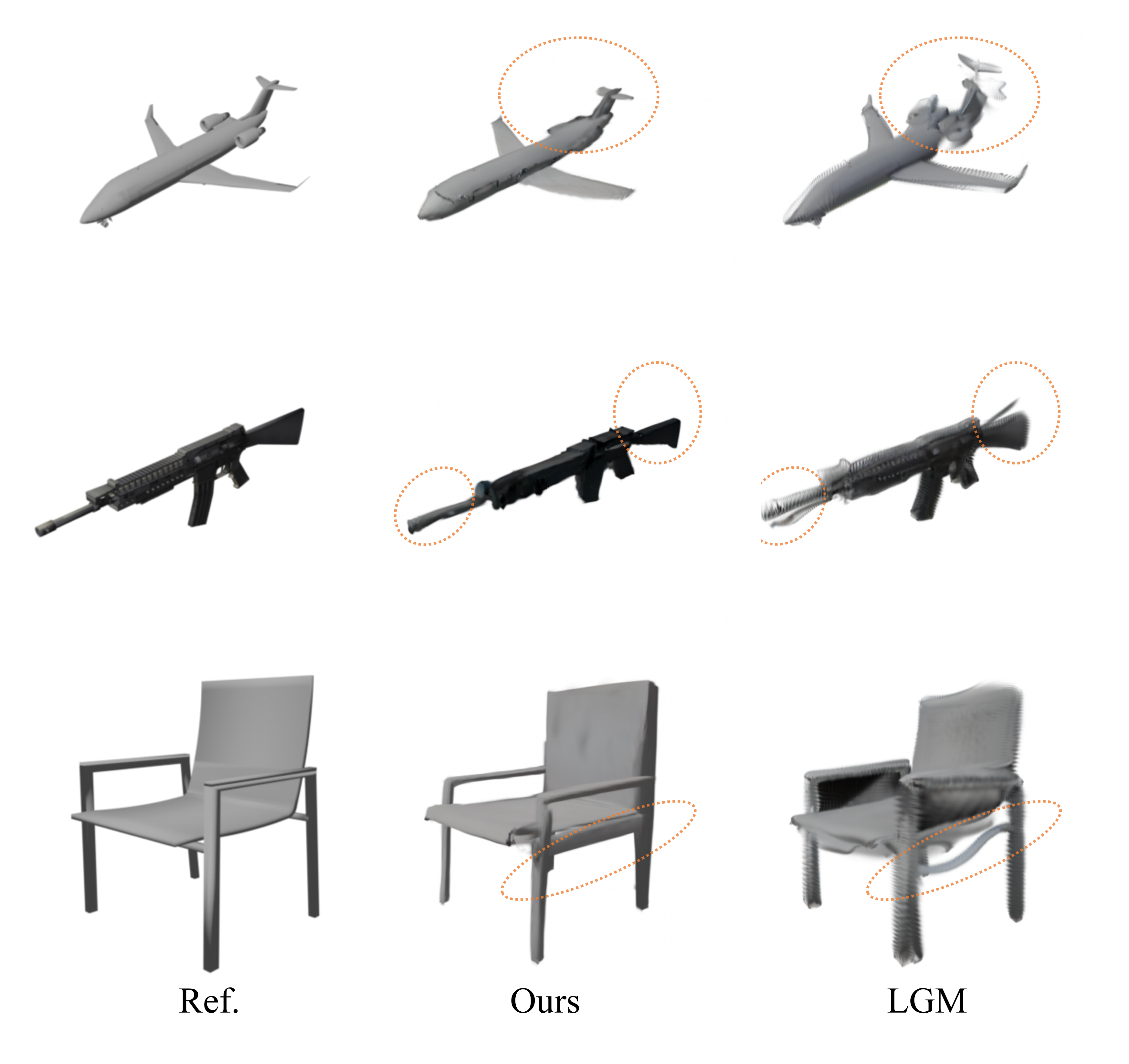}
    \caption{Geometry Comparison with  LGM \cite{tangLGMLargeMultiView2024}. Our method demontra better geometry and show less artifacts.}
    \label{fig:cmp_geometry_LGM}
\end{figure}

We present our Qualitative results in \cref{fig:render_cmp}.
LAS-Diffusion fails to capture the geometry present in the 2D sketch. This is due to the domain gap between synthetic 2D sketches and free-hand 2D sketches, which highlights its weak generalization ability. Additionally, it cannot produce texture or color since it adopts an SDF-based representation. 
For LGM \cite{tangLGMLargeMultiView2024}, which adopts an image-based approach, the result is more similar to the ground truth reference image. However, since it infers the multi-view from only a single view, the result is ambiguous. As a result, it produces a fused object that suffers from inconsistencies (e.g., the plane and gun in \cref{fig:cmp_geometry_LGM}) and illusory artifacts (e.g., the chair in \cref{fig:cmp_geometry_LGM}). In contrast, since our method adopts a native 3D representation, it shows better geometry consistency and fewer artifacts.

The quantitative results are shown in \cref{tab:3dgen}. Our method achieves about 50\% lower CD than LGM, demonstrating better geometry generation. We believe this is due to the use of a sketch condition, resulting in better alignment with the ground truth (GT). For visual metrics, we achieve better FID and CLIP scores, despite our method not using direct image conditions. 

We also tested the average inference time, which shows that our method is slightly faster than LGM, thanks to the 3D native representation and the use of a single-stage denoising process during inference.

\begin{table}[htbp]
    \centering
    \resizebox{0.9\linewidth}{!}{
    \begin{tabular}{l|cccc}
        \toprule
        Method & CD $\downarrow$&  FID$\downarrow$ &  CLIP score $\uparrow$ & latency (s) $\downarrow$\\
        \midrule
        LGM \cite{tangLGMLargeMultiView2024} & 0.0019 &	93.66 &	0.765  & 1.634 \\
        Ours &   \textbf{0.0008}&	\textbf{51.84}&	\textbf{0.826}  &  \textbf{ 1.460}\\
        \bottomrule
    \end{tabular}
    }
    \caption{Quantitative evaluation of 3D geometry and visual quality.}
    \label{tab:3dgen}
\end{table}



\section{Conclution}
In conclusion, we introduced VRSketch2Gaussian, a VR sketch-guided framework for efficient, multi-modal 3D object generation using 3D Gaussian Splatting. We also present VRSS, the first paired dataset combining VR sketches, text, images, and 3DGS. Our method leverages multi-modal conditioning and high-fidelity native 3D generation for fast, texture-rich 3D object synthesis. Experiments on VRSS demonstrate high-quality results in both geometry and visual fidelity. We believe our contributions will advance the field of 3D generation and ultimately find application in VR creation tools.

{
    \small
    \bibliographystyle{ieeenat_fullname}
    \bibliography{VRsketch2Gaussian}
}


\end{document}